\documentclass[letterpaper, 10 pt, conference]{ieeeconf}  

\IEEEoverridecommandlockouts    

\usepackage{graphicx}
\graphicspath{{figures/}}

\usepackage{amsmath} 
\usepackage{amssymb}  
\usepackage{balance}
\usepackage{booktabs}

\usepackage[separate-uncertainty = true]{siunitx}
\usepackage[hidelinks]{hyperref}
\usepackage[capitalise]{cleveref}

\usepackage{graphicx}
\usepackage{svg}

\Crefname{figure}{Fig.}{Fig.}
\Crefname{section}{Sec.}{Sec.}

\title{\LARGE \bf
Sensorized Soft Skin for Dexterous Robotic Hands
}

\author{Jana Egli$^{1}$, Benedek Forrai$^{1}$, Thomas Buchner$^{1}$, Jiangtao Su$^{2}$, Xiaodong Chen$^{2}$, and Robert K. Katzschmann$^{*1}$
\thanks{$^{1}$Soft Robotics Lab, IRIS, D-MAVT, ETH Zurich, Switzerland
{\texttt{\{eglij, tbuchner, rkk\}@ethz.ch}}}%
\thanks{$^{2}$School of Materials Science and Engineering, Nanyang Technological University, Singapore
        {\texttt{\{jiangtao001,chenxd\}@ntu.edu.sg}}}%
\thanks{$*$ Corresponding author: \href{mailto:rkk@ethz.ch}{\tt rkk@ethz.ch}}
}

\begin{document}

\maketitle
\thispagestyle{empty}
\pagestyle{empty}

\begin{abstract}
Conventional industrial robots often use two-fingered grippers or suction cups to manipulate objects or interact with the world. 
Because of their simplified design, they are unable to reproduce the dexterity of human hands when manipulating a wide range of objects.
While the control of humanoid hands evolved greatly, hardware platforms still lack capabilities, particularly in tactile sensing and providing soft contact surfaces. 
\textcolor{black}{In this work, we present a method that equips the skeleton of a tendon-driven humanoid hand with a soft and sensorized tactile skin. Multi-material 3D printing allows us to iteratively approach a cast skin design which preserves the robot's dexterity in terms of range of motion and speed. We demonstrate that a soft skin enables firmer grasps and piezoresistive sensor integration enhances the hand's tactile sensing capabilities.}
\end{abstract}

\section{INTRODUCTION}

The complexity of human hands allows us to grasp almost any object. Thanks to our skin with its integrated haptic sensors, we can even perceive items in the hand despite imperfect visual feedback. The high density of nerves and sensors in the fingers and palm helps us to judge the success of a grasp.

Visual feedback helps us a lot in daily object manipulation. However, we also integrate touch perception~\cite{cockburn_multimodal_2005}. Today, robots can already approximate human-level manipulation skills for narrow tasks with visual feedback like~\cite{handa2023dextreme}, but tactile sensor information has been shown to enhance, or, in special cases, even substitute visual information~\cite{yin2023rotating}.


In contrast to humanoid hands, robotic grippers tend to be either strong or precise~\cite{samadikhoshkho_brief_2019,billard_trends_2019}, but lack their structural complexity and dexterity.
Robotics researchers have been working on reproducing the structure of the human hand in robots since the early 80's~\cite{salisbury1982articulated,liu1999new,dai2009orientation,toshimitsu_getting_2023}. These works show similar degrees of freedom as a human hand, but they have no tactile sensing. This lack in tactile sensing limits the proprioceptive capabilities and therefore impacts dexterity and versatility of these humanoid robotic hands.

Tactile sensing is just one of several features of human hands where current robotic hands fall behind human hands. Current robotic grippers are mostly made from stiff and rigid structures~\cite{kadalagere_sampath_review_2023}. The human hand, in contrast, is constituted of flesh and skin around a load-bearing skeleton. These soft properties allow us to perform conforming grasps of objects and manipulate delicate objects without precise joint encoders. Our skin also increases the contact area and friction with the object, therefore enables manipulation of heavier loads.

   \begin{figure}[t]
      \centering
      \includegraphics[width=\linewidth]{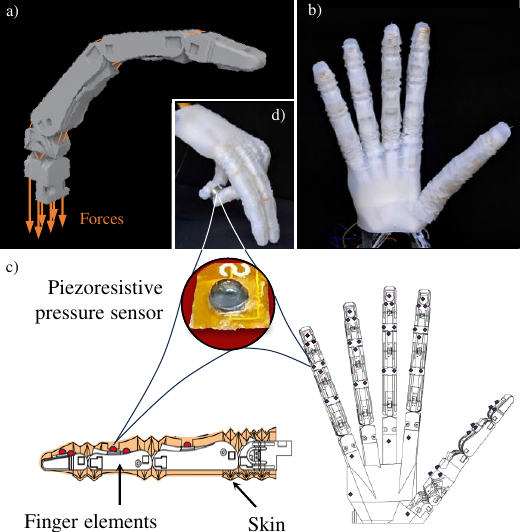}
      \caption{a) We used the Faive hand~\cite{toshimitsu_getting_2023} as a robotic platform for this work. The biomimetic, tendon-driven hand is equipped with rolling contact joints. b) We cast a silicone skin with origami-inspired structures to enhance the gripper's manipulability and buckle-free bending during hand movements. c) To enable tactile feedback from the hand, we attached 46 individual piezoresistive pressure sensors to the hand's fingers and palm. d) A tSNE algorithm distinguished different objects held by the sensorized gripper.}
      \label{fig:overview}
   \end{figure}

Accordingly, enveloping robotic hands with soft skins could potentially lead to many advantages. Hands with skins have a longer lifetime due to joint protection. Dust and objects won't protrude into the joint and water is repelled. On the mechanical side, larger contact surfaces and friction can make the manipulation of objects easier~\cite{mohammadi_practical_2020}. 

However, designing a skin for a biomimetic robotic hand that is not limited to simple geometries and can be manufactured fast remains a challenge. Artificial soft skins come with a wide range of issues such as  protrusion of skin material into the flexing joint region, which can limit the range of motion and induce additional demand in power supplied to the joint motors~\cite{tavakoli_anthropomorphic_2017,mohammadi_practical_2020}. These prior works~\cite{tavakoli_anthropomorphic_2017,mohammadi_practical_2020}, however, chose simple hinge joints instead of bioinspired joint designs. In contrast, the Faive robotic hand used in this work has biomimetic rolling contact joints~\cite{toshimitsu_getting_2023} and allows for finger abduction. 

To enable seamless actuation of these dexterous hands, a new, more complex skin design was needed.
We chose a 3D-printed origami structure for our skin as such configurations have been shown to follow deforming forces and restore initial configurations~\cite{zhang_3d_2021,tao_4d_2020}. The printing of fine origami structures, however, constrains the flexibility in design. The printer's capabilities can limit the structure resolution~\cite{walker_additive_2019}. The integrity of the prints is higher when the skin's structure is thicker, but this increased thickness also leads to reduced folding ability~\cite{bhat_composite_2023}.

We introduced a method to rapidly prototype soft structures using a multi-material printer for the design of complex skin geometries. To ensure the functionality and integrity of the skins, especially the origami parts, we then cast the final version of the skin. Once cast, we embedded custom-made tactile sensors into the skin structure. Reliably and repeatably applying tactile sensors on robotic hands is hereby particularly challenging. The placing of piezoresistive sensors beneath soft skin can help to detect objects grasped with the robotic hand and make statements about the grip strength. 

In this work, our contributions are the following:
\begin{itemize}
    \item We present a method for fast prototyping and optimizing of a soft skin for our dexterous robot hand, the Faive hand~\cite{toshimitsu_getting_2023}. We hereby use multi-material 3D printing for the rapid evaluation of design parameters for a soft skin with complex structure.
    \item We evaluate our skin's efficiency in grasping objects with the uncovered skin-free robotic hand through quasi-static pulling tests. We also assess the skin's impact on the robotic hand in dynamic scenarios by benchmarking the Faive hand's responses to alternating commands of up to 2.5 Hz with the skin on and off.
    \item We mount custom-made piezoresistive pressure sensors at key contact areas of the robotic hand, and present a classification method to mimic the proprioception of human hands.
\end{itemize}

\section{METHODS}
\subsection{Design}

We designed a 1mm thick skin around the Faive robotic hand~\cite{toshimitsu_getting_2023} in a CAD software. Besides a full-hand skin, we did separate designs for finger and palm. Previous work reported an ideal skin thickness of 1mm to trade deformation and fabrication limits~\cite{tao_4d_2020,bhat_composite_2023,dalaq_experimentally-validated_2022}. The index and ring finger were identical. A hexagonal origami skin spanned the finger joints. Soft origami gripper have already shown to be able to grasp objects of irregular shape. Their compliance helps in folding around items~\cite{liu_3d-printable_2021,chen_soft_2021}. 

\begin{figure}[thpb]
      \centering
      \includegraphics[width=\linewidth]{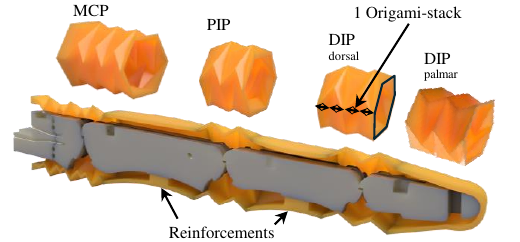}
      \caption{\textcolor{black}{Cross section of the index finger design. The three joints - MCP, PIP and DIP - and palmar skin reinforcements are highlighted. The reinforcements serve to direct the skin motion during finger flexion to avoid buckling, which can restrict the joints' range of motion. The DIP joint is shown from both, the palmar and dorsal side. The palmar side is pleated in shape whereas the dorsal side has a twisted, hexagonal origami structure. These structures allow the skin to distribute the deformation forces in the skin during finger motion.}}
      \label{fig:scheme}
   \end{figure}

Our Metacarpophalangeal joints (MCP) had a circular symmetric hexagonal base. The profile of the base was rotated by 30 degrees from one origami plane to the next. The MCP joints consisted of six stacked origamis for the Index, Middle, Ring and Pinky finger. The Proximal Interphalangeal (PIP) and Distal Interphalangeal (DIP) joints were hexagonal and symmetric only with respect to the sagittal plane. The skins of these two joint types consisted of four stacked origamis (see Figure~\ref{fig:scheme}). The Carpometacarpal (CMC) joint of the thumb was hexagonal and circular symmetric with eight stacked origamis. The thumb's MCP joint skin was a sagittal plane symmetric stack of four origamis. The PIP joint of the thumb was also symmetric w.r.t the sagittal plane and designed as a stack of four origamis. 

Adjacent joints were connected by skin, surrounding the phalangeals. For the four fingers, distal and proximal phalangeal skin parts had a reinforced, \textit{i.e.}, thickened inner part on the palmar side of the skin. The thickness was determined as the distance between the skin and the robotic skeleton.
We designed the palm skin as a smoothly shaped cover of the skeleton.

\subsection{Fabrication}

We did the prototyping using a multi-material 3D-Printer. The skin made of custom made SEBS with shore hardness 18A and AquaSys120 (Infinite Material Solutions Inc.) as support structure. The water-soluble support allowed us to remove it from the skin without tearing the soft structure. The drawback of 3D-Printing with multi-material printers is, that it's slow compared to conventional 3D printer because of the time consumed for switching the print-heads and for heating up the materials to their extrusion temperature for each layer. Printing soft materials is not as reliable as casting in terms of structure integrity. Printed layers tended to separate and printing resolution is limited by the nozzle size and expansion of the soft material after leaving the nozzle. This limitations challenged the printability of origami structures.

\begin{figure}[thpb]
      \centering
      \includegraphics[width=\linewidth]{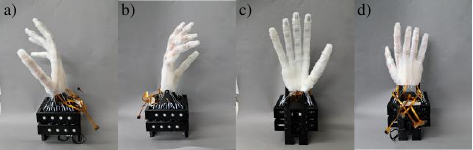} 
      \caption{\textcolor{black}{A silicone skin was cast in one piece and mounted on the Faive hand. Lateral views in a) and b). Palmar side in c) and dorsal side in d). Tactile sensors on flex PCBs are placed on the hand skeleton and wrapped by the silicone skin.}}
      \label{fig:glove}
   \end{figure}

Casting is suitable for accurate fabrication of soft designs. Other works discussed the downside of casting being even more time-consuming~\cite{hernandez_current_2023,walker_14_2019}. First, a mold has to be 3D-printed with a rigid material such as PLA before pouring the silicone into it. Then the material needs to cure before the PLA mold can be removed from the silicone. 
We first showed that the dynamic behavior such as range of motion (ROM) and latency of a single finger design is not negatively affected by the skin, before we decided to cast our final skin design in form of a complete skin (see Figure~\ref{fig:glove}). Therefore, the skin will be smoother and more robust.
We took the negative of the skin to generate molds for the casting of silicone skins. To cast the full-hand skin, fingers, and palm, we split the molds into parts, which could be assembled around the inner parts of the hand's skin. Each finger mold consisted of two parts which were aligned in the sagittal plane. The palm mold was separated into four pieces to minimize the number of parts.
We chose the directions in where the mold parts will be joined such that easy mold assembly and skin removal after casting the silicone of hardness Shore A10 (DragonSkin 10, Smooth-On, Inc.) is ensured. We introduced isopropanol between silicone and PLA to separate the mold and the skin.

\subsection{Sensors} 

We functionalized the skin with piezoresistive pressure sensors. Their placement density aligns with the biologic sensitivity found in literature~\cite{gardner_simulation_1990,vallbo_properties_1984}. Three on the finger tips, three on the distal phalangeals, two on the proximal phalangelas and five at the edges of the robotic skeleton's palm .

The sensors consisted of a piezoresistive sensing layer, which are covered with a silicone hemisphere. The silicone tips were glued on with nonconductive epoxy polymer. The silicone transfers the force over the piezoresistive sensing layer to the electrodes (see Figure~\ref{fig:sensors2}). By compression of the conductive material the resistance is reduced when force is applied. The sensors capability to detect different amount of forces could be shown in previous research~\cite{cui_freestanding_2022}. The sensors were serially connected. For each finger and the palm a separate flexible PCB is placed between the robotic skeleton and the skin. The  flexible PCBs are routed behind the hand for the read out.

\subsection{Testing} 

To assess the effect of our custom skin on the manipulation capabilities of the Faive Hand, we performed both dynamic and static tests.

For the dynamic performance evaluation, we compared printed and cast finger skin prototypes without sensors. We commanded flexion of the PIP and MCP joint and abduction (ABD) of the MCP joint. The inputs were step and sinusoidal commands of frequencies ranging from 0.5Hz to 2.5Hz in 0.5Hz increments. 
Both the cast skin and individual fingers of the final printed version were tested against the hand without skin (see Figure~\ref{fig:rom}). We calculated the command gain as the maximal ROM over the achieved ROM. For the latency, we calculated the delay time between commanded finger position and reached position. 

\begin{figure}[ht]
      \centering
      \includegraphics[width =\linewidth]{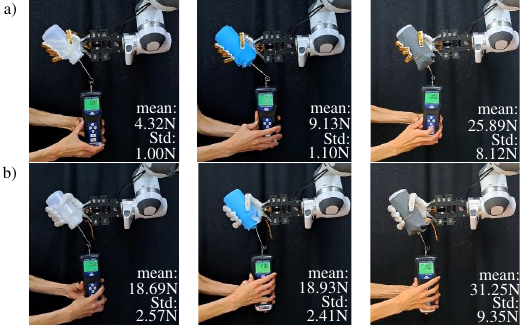}
      \caption{The ability to hold an LDPE bottle in a power grasp without (a) and with skin (b) was tested for the different rough surfaces. The hand held an uncovered LDPE bottle and a bottle wrapped in sandpaper and a lab glove respectively. We pulled at the bottles with a quasi-static force and increased the force stepwise until slippage of the bottle. Mean and standard deviation of the resisted forces over five trials are displayed for each hand-object combination.}
      \label{fig:ForceTest}
   \end{figure}

To assess the skin's impact in static scenarios, we conducted a quasi-static pull test on the grip strength. We used an LDPE bottle as the benchmark object with and without change of surface. We attached a hook to the bottle's cap, and through it, we pulled the bottle in a quasi-static manner along its longitudinal axis. We logged the pulling force with the help of a force gauge attached directly to the hook. The grasp type was a power grip commanded with 500mA maximal motor current. To control the friction between the hand and the bottle, we added sandpaper with a grit size of 600 and lab gloves wrapped around the bottle in two other experiments (see Figure~\ref{fig:ForceTest}). \textcolor{black}{Each surface test was repeated five times for both, skin and no skin setup. We calculated the mean and standard deviation of the maximal force that could be resisted in the five tests per experiment.}

We characterized the sensors before integrating them in the skin. We analyzed the sensor response to increasing force application. Also the change in resistance over time when pressing the silicone tip with 1.52N and reducing the force to 0.45N and the drift of resistance over 5000 was recorded. In the cycling test, We alternated between forces of 0.8N and 0.2N (see Figure \ref{fig:sensors2}). \textcolor{black}{Before the sensor integration on the hand we compared the signals from flexible PCBs placed flat on a table and wrapped around table edges to test the influence of bending on the sensing.}

For the sensor evaluation in use with the robotic hand, we mounted them on the PLA skeleton of the Faive hand. The flexible PCBs were covered with the cast silicone skin. Mounting the skin on the sensorized hand had to be done with caution to avoid displacement of the sensor's silicone tips.

The sensor performance test consisted of grasping a set of objects and interaction. We grasped a 50g weight between thumb, index and middle finger and a mustard can with a power grip. As human-robot interaction task, we did a handshake between robotic and human hand. The relaxed hand position and a closure of the hand without holding any object were the comparisons. The sensitivity of the sensors was determined as change in resistance during the interaction (see Figure \ref{fig:sensors2}).

\subsection{Object detection with grasping}

To show the potential of the embedded tactile sensors, we implemented a simple object identification based on their resistance values. Compared to recent successful tactile systems like~\cite{yin2023rotating}, we had abundant data: each of the 46 sensors has been monitored through an ADC at a rate of 20Hz. This was first fed through a median filter of width 0.5 seconds to get rid of outliers that might have appeared in the case of hand motions or unreliable sensor contact states. Similarly to the object classification approach from B. S. Homberg et al.~\cite{homberg2015haptic}, we use unsupervised learning to cluster the filtered data stream from the grasped objects. To accommodate the relatively high number of dimensions, we chose to visualize the incoming measurements with t-distributed Stochastic Neighbor Embedding~\cite{van2008visualizing}. \textcolor{black}{The test was performed with and without skin.}

\section{RESULTS} 

\subsection{Skin Quality} 

The layers of printed skins were visible and susceptible to tears. Removal of support and mounting of the skin on the hand led to ruptures. Narrow origami structures and conversion zones from origami to phalangeal parts often showed discontinuities of the skin.

cast skins had smooth surfaces with only few air bubbles. Tears from mold removal were merely prominent at the MCP joints and locations where mold parts have been assembled. Mold release spray on PLA prior to silicone casting as well as rinsing of isopropanol over the silicone skin after curing facilitated the separation of mold and skin. The occurrence of skin tears was thereby reduced.

\begin{figure}[thpb]
      \centering
      \includegraphics{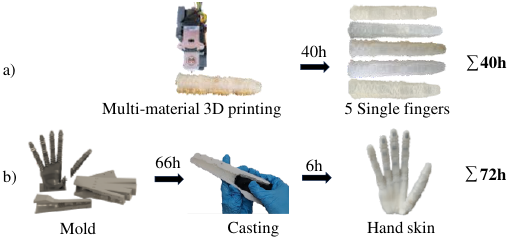}
      \caption{\textcolor{black}{a) The process of multi-material 3D printing allows fast prototyping of finger structures and uses only a single step. SEBS printing is fast but the resulting skin structure is less robust. b) Molding is a time-consuming method to create skins in one piece with an even surface and large structural integrity. We carry out molding in two steps: Fist, we print molds from PLA with a simple FDM printer, then we cast DragonSkin silicone in the mold, wait for it to cure and demold. Casting takes 1.8 times longer than printing but the silicone skins are of higher quality than SEBS skins.}}
      \label{fig:manufacturing}
   \end{figure}

Next to the integrity of the skin, the time needed to fabricate the skin determined the choice of manufacturing method. Printing all five fingers is in total 1.8 times (around 32h) faster than casting an entire skin around the hand. The estimated printing time was calculated automatically by the printers slicing program. For the multi-material printer, the time might exceed the estimate by two to four hours depending on the amount of layers. The additional time was due to the heating-up of the nozzle to the materials extrusion temperature and the print head switch (see Figure~\ref{fig:manufacturing}).

Fabricating the final skin took 72 hours when including the time needed for the mold printing. Once the mold is ready, casting the skin is a more time efficient and reliable manufacturing method than printing.

\subsection{Performance}
Dynamic tests showed minor difference between casting and printing of single fingers as well as without skin. The latency for printed finger skin application was only 0.5s at 2.5Hz compared to responsiveness of hands without skin. And zero for cast finger skins.
The range of motion of finger joints was reduced by approximately 0.05rad at 2.5Hz input for finger skins and without skin.

\begin{table}[ht]
\begin{center}
\caption{Difference in range of motion in degrees of the hand with and without skin.}
\begin{tabular}{lrrr}
\toprule
\textbf{Model} & \textbf{2.5Hz} & \textbf{1.5Hz} & \textbf{0.5Hz} \\
\midrule
$\Delta$ Thumb DIP & 16.44 & 13.01 & 5.44\\
$\Delta$ Thumb ABD & 0.29 & 5.43 & 0.88\\
$\Delta$ Thumb MCP & 1.70 & 6.85 & 1.93\\
$\Delta$ Thumb PIP & 2.02 & 5.44 & 2.60\\
$\Delta$ Index ABD & 0.26 & 1.50 & 0.14\\
$\Delta$ Index MCP & -4.86 & -10.24 & -2.04\\
$\Delta$ Index PIP & 0.49 & 6.47 & 2.85\\
$\Delta$ Middle ABD & 0.03 & 1.44 & 0.15\\
$\Delta$ Middle MCP & -0.11 & 3.78 & -0.03\\
$\Delta$ Middle PIP & 0.01 & 1.74 & -0.92\\
$\Delta$ Ring ABD & 0.13 & 1.63 & 0.39\\
$\Delta$ Ring MCP & -3.53 & -3.60 & -3.55\\
$\Delta$ Ring PIP & -1.06 & -2.34 & 1.72\\
$\Delta$ Pinky ABD & -0.63 & -4.01 & -1.22\\
$\Delta$ Pinky MCP & 1.82 & 3.17 & 3.20\\
$\Delta$ Pinky PIP & -0.54 & 1.15 & 3.14\\
\bottomrule
\end{tabular}
\end{center}
\label{tab:ROM}
\end{table}

\begin{figure}[ht]
      \centering
      \includegraphics[width=\linewidth]{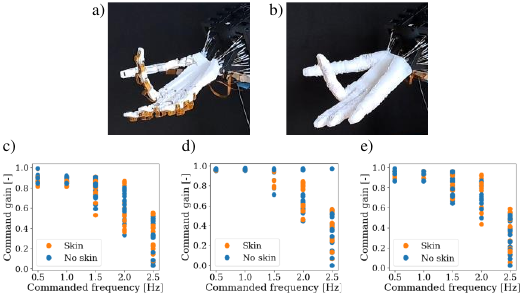}
      \caption{a) and b) Dynamic evaluation of the robotic hand without and with the skin. After running step and sine responses on each joint of each finger, we conclude that there is no detrimental difference in range of motion with or without the skin for the c) ABD, d) MCP and the e) PIP joint. \textcolor{black}{It can also be seen that the skin's effect on the range of motion is even smaller for higher commanded motion frequencies. We would hypothesize that at higher frequencies, the actuator's dynamics are slow enough to make a larger impact on the motion range than the presence of the skin.}}
      \label{fig:rom}
   \end{figure}

In the comparative tests of cast skin and no skin, we showed that both setups perform equally well  (see supplementary video). The range of motion for all joints (PIP flexion, MCP flexion and abduction) was maintained with the skin on the hand. At high frequencies (2.5Hz), the range of motion was reduced for skin and no skin hands. The difference in ROM between skin and no skin tests can be taken from figure \Cref{fig:rom}
and \Cref{tab:ROM}. Since we used the EKF of the hand for our range measurements, our values were susceptible to calibration errors, which sometimes resulted in negative values - like in the case of the index MCP joint. Excluding these cases, we saw that the maximal impacts are in the range of 10 degrees, so we concluded that there was no major adverse effect from the skin on the dynamic performance.

We saw in force tests, that a power grasp could hold a bottle of smooth surface (LDPE and lab glove/rubber) against higher force than without skin (see supplementary video). The mean force was quadrupled for LDPE and doubled for rubber.
Surfaces with high friction coefficients allowed for higher forces in both cases, with an increased standard deviation in maximal force.

\begin{figure}[ht]
      \centering
      \includegraphics[width =\linewidth]{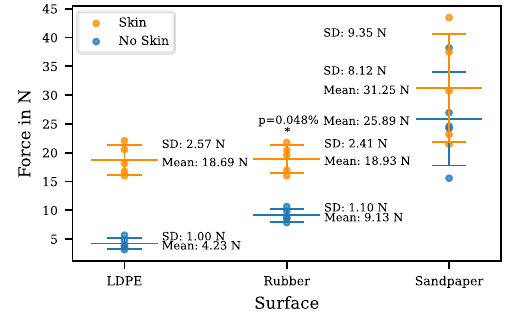}
      \caption{\textcolor{black}{The hand with skin resisted higher pulling forces than without skin for all the different objects' surfaces. At higher pulling forces, the data from five trials per hand-object combination is more spread. Objects with rubber surfaces can be held significantly better with a skin with a p-value of 0.048\%.}}
      \label{fig:ForceStats}
   \end{figure}

A paired T-test revealed a significant difference in resistance to pulling force between hands with and without skin when the grasped object had rubber surfaces. The p-value was with 0.048\% under the 5\% significance level (see Figure~\ref{fig:ForceStats}). Rough surfaces like sandpaper could be held comparably well for hands with and without skin.

\begin{figure}[ht]
      \centering
      \includegraphics[width=\linewidth]{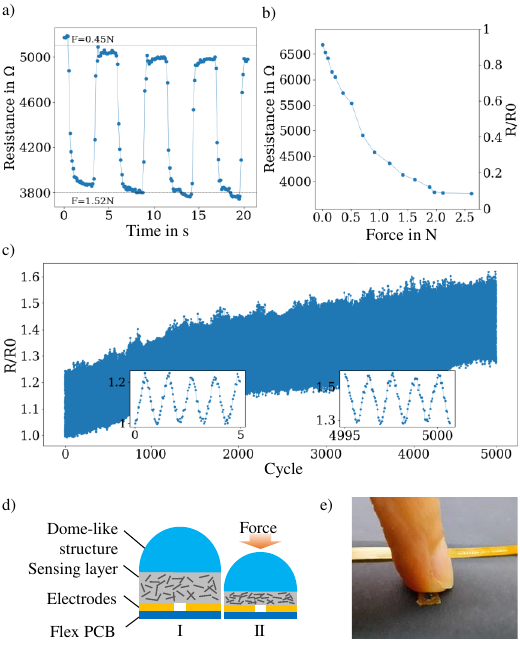}
      \caption{Tactile sensor setup. a)~Four cycles of forces in the range of \qtyrange{0.45}{1.52}{\newton} applied to a sensor over \qty{20}{\second}. b)~Sensor characterization showing the relation between the sensor's resistance and force applied to the sensor. c)~The cyclic loading of the piezoresistive pressure sensors over 5000 cycles shows some drift. The baseline resistance and force-response resistance change. The absolute difference in resistance for certain forces remains in the same range for early and late cycles. d)~The piezoresistive pressure sensors are comprised of a substrate (flex PCB), two electrodes separated by a small gap and the sensing layer that shows a change in resistance between uncompressed (I) and compressed (II) states. A dome-like structure that distributes forces applied to the sensor completes the sensor setup. e)~The miniature piezoresistive pressure sensor activated by an index finger.}
      \label{fig:sensors2}
   \end{figure}

We characterized the sensors to have a nonlinear relationship between applied force and resistance. At 2.5N, the relative resistance is reduced by approximately 40 percent. With prolonged cycling force application, the resistance increases for untouched and pressed status. The drift is below 1000 Ohm for 5000 repetitions, which is a relative change of 0.3 (see Figure~\ref{fig:sensors2}).

\begin{figure}[ht]
      \centering
       \includegraphics[width =\linewidth]{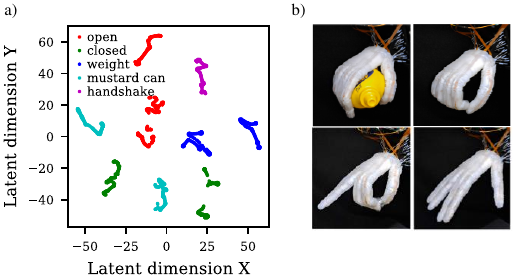}
      \caption{Using t-distributed Stochastic Neighbor Embedding (t-SNE), we embedded the 46-dimensional sensor readings in a 2 dimensional latent space, arriving to separate clusters after 30 minutes of consecutive experiments of grasping different objects (a). To show the extra performance that tactile sensing provides to simple joint tracking approaches, we also recorded data with an empty grasping hand, along with different household objects (b).}
      \label{fig:tsne}
   \end{figure}

In the evaluation of sensor responses in grasping and interaction tasks, we saw that there is a characteristic pattern of responding sensors for different objects and positions. A t-SNE analysis allowed us to distinguish between the open and closed hand position as well as a handshake, mustard can and 50g weight grasp. \textcolor{black}{The sensor responses can be interpreted as a characteristic of the object shapes. The same objects could be distinguished with and without skin.} From Figure \ref{fig:tsne}, we conclude that the combination of sensors which change their resistance in a task is specific to that grasp task. The task clusters are not overlapping. However, some sensors tended to become unstable with time. Due to the custom-made sensor adhesion on the flexible PCBs, some piezoresistive platelets came loose, especially when putting the skin on or rigorous hand motions. Loosening led to unsteady resistance signals. Sensor stripes of individual fingers could be exchanged to reestablish the sensitivity. The fabrication technique of the sensor stripes also influenced their individual range of resistance change. 

\section{DISCUSSION}

In our study, we have shown that enveloping dexterous robotic hands with a soft skin enhances the grasp quality while keeping our robotic hand's kinematic and dynamic properties intact. So far, no method has been established that does not sacrifice the flexing abilities of joints or the feasibility of manufacturing geometrically complex skin designs. Both problems have been reported by Mohammadi~\cite{mohammadi_practical_2020} and Tavakoli~\cite{tavakoli_anthropomorphic_2017}. Origami structures, which meet the dynamic demands, are hard to parameterize quickly and their shape is a challenge in fabrication.
Using multi-material 3D printing with SEBS 18A and PLA allowed us to find optimal design parameters, which fulfill the dynamic demands of skins. Because the layers of printed origami skins are prone to rupture~\cite{roels_additive_2020} and have a poor appearance, final designs are preferably being cast. Although molding took around 32 hours longer than printing, the molded outcomes are more robust and suitable for many grasp operations. Molding also allowed us to join the finger skins with a palm skin.

We showed that cast skins enhance the grasp performance of smooth objects with a power grasp. Compared to robotic hands without a skin, soft skins enable holding items with a low frictional surface, like LDPE or rubber, against up to 4 times more force. For more reliable statements about significance, a larger data set would be needed. 
We also had to take into account that the skin adds additional payload to the motors that are powering the joints. For a firm grasp, low motor currents consume energy which is caused to a large amount by the skin which tries to pull the fingers back into an extended position. The range of motion could be conserved by applying high enough currents for joint flexion and by designing the joint´s skin as hexagonal origami. \textcolor{black}{The largest reduction in range of motion was seen for the thumb's DIP joint. We assume this is due to too few strength of the motors to bend the joint against the skin.} The same dynamic performance as without skin can be maintained. This is of special importance when handling objects. We validated, that the latency between sinusoidal and step input commands is in the same range with and without skin. The skin is especially also not slower than the version without skin at high frequency commands of 2.5Hz. 

We placed piezoresistive pressure sensors with a hemispheric silicone tip between the robotic skeleton and skin. We could distinguish grasps of differently sized objects like mustard can and a 50g weight. The set of sensors that are distributed over the hand is also indicative of the hand's state. Relaxed open and closed finger positions were differentiable from the object grasps and human-robot interactions like handshaking. \textcolor{black}{The skin did not impair the detectability of various objects but rather helped in holding them with more stability. We saw that the silicone allows to transfer the forces over the silicone to the sensors.} Because custom-made sensors tend to dislocate with time, their sensitivity might become compromised in prolonged dynamic use. \textcolor{black}{We used only 46 sensors. Therefore, the sensor density across the hand is not very high. It would be beneficial for object classifications to have a better spatial resolution. More sensors would help in localizing point contacts with objects on a smaller scale. The sensor nonlinear drift is relatively large but might stabilize after 5000 cycles of loading. Further calibration of the sensors is needed for prolonged grasping tasks.}

Supplementing robotic hands with tactile intelligence embedded between robot and a soft skin can contribute to making the hands more natural. This additional proprioceptive dimension can, in the future, be used to get tactile feedback and help in robotic learning-based control schemes. For this, hardware- and software-based questions need to be solved. For example, how can we fixate sensors best to avoid their dislocation during interaction? What increases our readout sensitivity and which machine learning tools lead to the best interpretation of the tactile feedback the skins provide?




\section*{ACKNOWLEDGMENT}
We are grateful for the ETH Zurich RobotX Research program funding and the SNSF Project grant \#200021\_215489.
We would like to thank Barnabas Gavin Cangan for supporting us with the measurement setup. We also want to thank Yves Haberthür for his work on the multi-material 3D printer.

\addtolength{\textheight}{-6 cm}   

\bibliographystyle{IEEEtran}
\bibliography{references_extra}

\end{document}